# Empowering Working Memory for Large Language Model Agents

Jing Guo, Nan Li\*, Jianchuan Qi, Hang Yang, Ruiqiao Li, Yuzhen Feng, Si Zhang, Ming Xu ╪

Tsinghua University

\*li-nan@tsinghua.edu.cn, ╪xu-ming@tsinghua.edu.cn

## Abstract

Large language models (LLMs) have achieved impressive linguistic capabilities. However, a key limitation persists in their lack of human-like memory faculties. LLMs exhibit constrained memory retention across sequential interactions, hindering complex reasoning. This paper explores the potential of applying cognitive psychology's working memory frameworks, to enhance LLM architecture. The limitations of traditional LLM memory designs are analyzed, including their isolation of distinct dialog episodes and lack of persistent memory links. To address this, an innovative model is proposed incorporating a centralized Working Memory Hub and Episodic Buffer access to retain memories across episodes. This architecture aims to provide greater continuity for nuanced contextual reasoning during intricate tasks and collaborative scenarios. While promising, further research is required into optimizing episodic memory encoding, storage, prioritization, retrieval, and security. Overall, this paper provides a strategic blueprint for developing LLM agents with more sophisticated, human-like memory capabilities, highlighting memory mechanisms as a vital frontier in artificial general intelligence.

## 1. Introduction

The development of large language models (LLMs) has marked a significant advancement in the field of artificial intelligence (AI), particularly in the realms of language understanding, generation, and reasoning. These models, exemplified by OpenAI's ChatGPT, are characterized by their extensive architectures and massive parameter sets, having been trained on vast corpora of text (Brown et al., 2020; Rillig et al., 2023). Despite their impressive linguistic capabilities, a critical challenge that persists is the effective management of memory to achieve more human-like intelligence. Cognitive psychology offers foundational frameworks, such as Baddeley's multi-component working memory model, which have been instrumental in understanding human memory (Baddeley, 2003). However, the application of these frameworks to AI architectures is not straightforward, and there are inherent limitations to how these human-centric concepts can be translated into artificial systems(Cornago et al., 2023).

Standard LLM agent designs lack robust episodic memory and continuity across distinct interactions. LLM agents typically have a constrained memory capacity, limited by the number of tokens they can process in a single exchange. This limitation restricts their ability to retain and utilize extensive context from previous interactions. Moreover, each interaction is treated as an isolated episode, with no linkage between sequential dialogues. This isolated, short-term memory hinders complex sequential reasoning and knowledge sharing in multi-agent systems. The absence of a robust episodic memory and continuity across interactions hampers the agents' ability to perform complex sequential reasoning tasks, which are essential for more advanced problem-solving capabilities (Figure 1). Especially in Multi-Agent Systems (MAS), the lack of cooperative communication among agents can lead to suboptimal outcomes. Ideally, agents should share immediate actions, or learning experiences to achieve common goals efficiently. To address these challenges, advancements in AI memory architectures, such as Neural Turing Machines and Memory Networks, have been proposed to enhance the memory capabilities of LLM agents (Graves et al., 2014; Sukhbaatar et al., 2015; Weston et al., 2015). These models aim to provide a more sophisticated framework for memory management, potentially enabling LLM agents to better mimic human-like intelligence and memory functions. However, these models often face challenges related to computational complexity, integration difficulties, limited generalization across tasks, dependency on extensive training data, and a lack of human-like flexibility and interpretability in their memory functions.

To address the above limitations, this paper delves into the development of working memory models, exploring their journey from the realm of cognitive psychology to their application in advanced LLM agents. Recognizing the challenges in adapting human working memory frameworks to artificial contexts, it highlights the shortcomings of traditional LLMs, particularly their lack of episodic memory and continuity in diverse interaction domains. To overcome these limitations, we propose a novel model that features a centralized Working Memory Hub and provides access to an Episodic Buffer. This design aims to endow agents with enhanced contextual retention and improved performance in intricate, sequential tasks and cooperative scenarios. The paper sets forth a forward-looking framework for crafting LLM agents with sophisticated, human-like memory functions and underscores the need for further research into optimizing the processes of memory encoding, consolidation, and retrieval to fulfill these objectives.

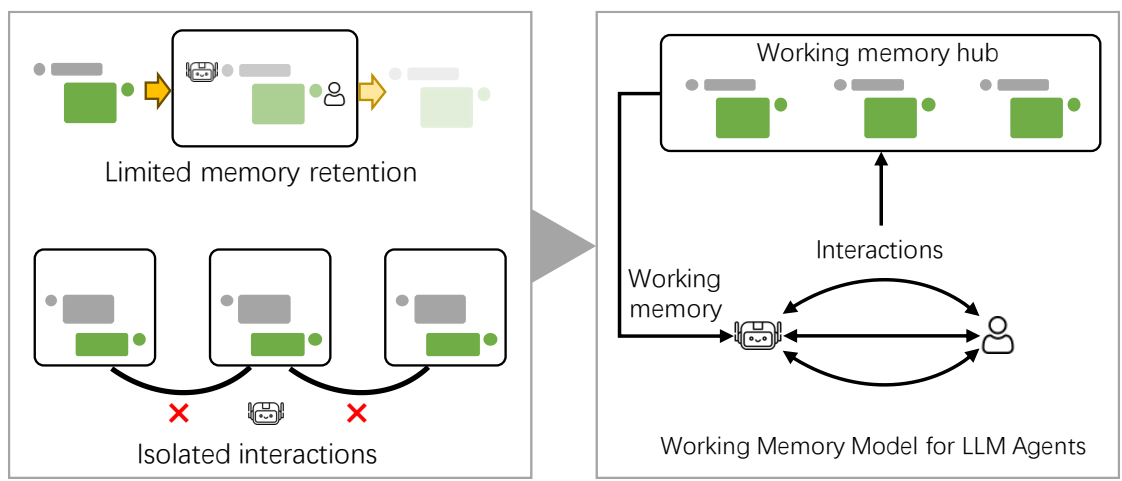

Figure 1 Solution for the shortcomings of traditional LLMs

## 2. Working Memory in Humans and LLMs

### 2.1 Working Memory Model in Cognitive Psychology

Emerging from the mid-20th century, cognitive psychology ushered in a transformative lens to the study of memory. Central to this paradigm shift was the "multi-component memory model" conceptualized by Atkinson and Shiffrin in 1968, segmenting memory into sensory memory, short-term memory, and long-term memory. Amidst these classifications, a burgeoning interest developed around the concept of working memory. Baddeley's introduction of the "working memory model" in 1974 delineated it not just as an alternate to short-term memory but as a nuanced, multi-component system dedicated to the transient storage and manipulation of information (Figure 2).

At the nucleus of this model is the **Central Executive**, acting as the supervisor. It orchestrates attention allocation, prioritizes information, and ensures effective rehearsal among its subsystems. This pivotal component communicates with two key subsystems. The **Visuospatial Sketchpad**, or the "inner eye", specializes in spatial and visual information. Whether it's visualizing landscapes or mapping routes, it's crucial. Moreover, it closely intertwines with our perceptual systems. For example, during reading, visual stimuli are rapidly encoded into this sketchpad, preparing the information for comprehension. The **Phonological Loop**, often termed the "inner voice", holds linguistic and phonological content in a speech-based format. Its storage, however, is fleeting unless the information is actively rehearsed. As the understanding of working memory evolved, the model was further refined. In 2000, Baddeley introduced the **Episodic Buffer** as an additional component, acting as a temporary storage that amalgamates information from varied sources, thus creating coherence among the Central Executive, Phonological Loop, and Visuospatial Sketchpad.

But the depth of this model goes beyond its components. A profound link binds working memory to long-term memory, with information shuttling between the two based on need and processing. From its origins, the working memory model has illuminated our grasp of cognition. Its principles have reverberated across fields, influencing educational methodologies aimed at boosting academic capabilities and inspiring the blueprint of intricate AI and computational architectures.

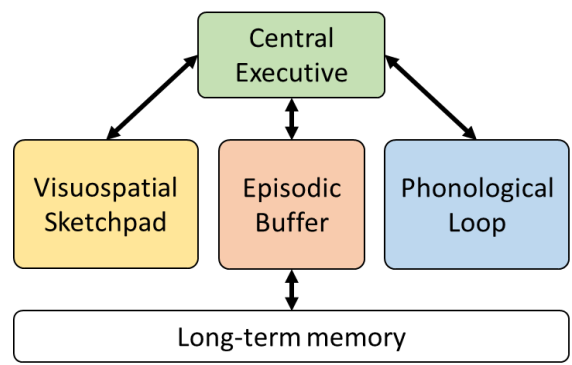

Figure 2 Working memory model in cognitive psychology

### 2.2 Working Memory Architecture in LLM Agents

Drawing inspiration from the human "working memory model", the LLM Agent's working memory offers a unique yet analogous architecture (Figure 3).

At the crux lies the **Central Processor**, which is, in essence, the LLM itself. It seamlessly intertwines the massive training data with real-time inputs, orchestrating the data flow and ensuring the appropriate information processing, analyzing, and making decision. **External Environment Sensor** stands as the gateway for the agent, facilitating real-time interactions with various external systems and databases. This ensures the agent's ability to get inputs and respond to dynamic external data seamlessly. The **Interaction History Window** is crucial for retaining short-term interactions. It preserves the recent lineage of interactions, offering contextual anchorage for the agent during ongoing tasks. However, due to the token limitations inherent in LLMs, the model can only retain a limited scope of the conversation. This constraint poses the **first challenge**, especially in extensive interactions, as the LLMs might lose older contextual information beyond its token capacity. Furthermore, the **second challenge** lies on the concept of '**task domains**' in the model, demarcated by the dotted lines. Each unique engagement or interaction session with an LLM is treated as its distinct domain. When a subsequent interaction is initiated, it essentially establishes a new domain, devoid of direct linkage to the previous ones. This organizational paradigm emerges from the way LLMs handle data streams, where every distinct interaction is processed as a separate episodic memory, isolated from preceding engagements. Thus, there is no episodic buffer as human "working memory model" includes.

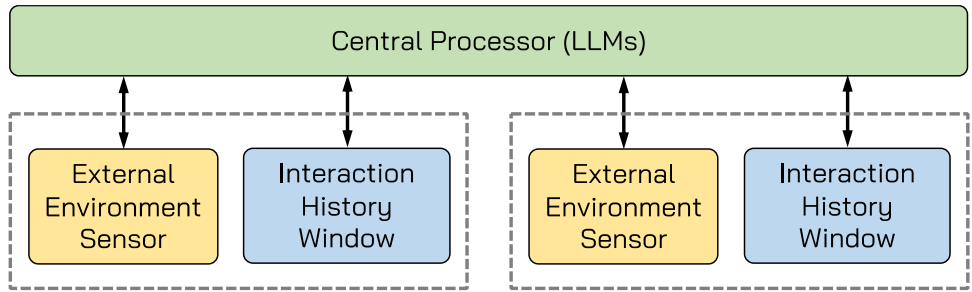

Figure 3 Current working memory model of LLM agents

The constraints of the current LLMs memory architecture become even more pronounced in multi-agent settings. As agents engage in collaborative tasks, the inability to directly share episodic memories across interaction domains severely limits collective progress. Agents lack mechanisms to consistently build upon or reference the experiences and knowledge gained by their peers over time. This isolated nature of individual agent memory prevents the emergence of a collective long-term memory spanning the shared interactions and goals of the agent group. Furthermore, the limited context retention within a single agent poses additional challenges when coordinating complex goals across multiple agents, where key information must be propagated across the system. Developing more interconnected and persistent memory capabilities is critical for enabling smoother collaboration and emergent cognition in multi-agent environments.

## 3. An Advanced Working Memory Model for LLM Agents

To address the above two challenges, current research endeavors are fervently exploring methodologies to aim for an even more robust LLM agent memory system. Addressing the first challenge, technologies have been developed to sustain long-term dialogues, effectively overcoming the context window limitations. Such advancements include recursively summarizing, RecurrentGPT, Long-term Memory(Q. Wang et al., 2023; W. Wang et al., 2023; Zhou et al., 2023). However, they did not change the working memory model. For the second challenge, transitioning from working memory to long-term memory, researchers aim to dissolve the "domain" separations between tasks. One way explored is the direct incorporation of memory within the model itself, leading to a personalized agent (K. Wang et al., 2023; Zhang et al., 2023; Zhu et al., 2023). While this approach fosters a more tailored agent, it comes with higher operational costs. Moreover, due to specific memory update mechanisms, there's a risk of losing intricate details from the working memory. To enhance the memory capabilities of LLM agents, an innovative memory architecture has been proposed, as depicted in the diagram (Figure 4). This evolved model incorporates additional components to address limitations of traditional working memory models.

**Working Memory Hub** unified hub orchestrates data flows between other components. It stores all inputs, outputs, and interaction histories to supply other elements like the Interaction History Window and Episodic Buffer. Similar to the above-mentioned working memory model, at the core is the **Central Processor** comprised of LLMs, acting as the brain of the agent. It ensures that information is processed, analyzed, and decisions are made based on a harmonious blend of historical and current inputs from the External Environment Interface. Also, **External Environment Interface** facilitates the continuous influx of dynamic external data into the system. It intakes real-time inputs from users and external sources and routes them to the Central Processor for analysis. Likewise, it captures the outputs from the Central Processor and disseminates them as responses. All inputs and outputs are stored in the Working Memory Hub. Drawing from the Working Memory Hub, **Interaction History Window** maintains a short-term cache of recent interaction history to provide contextual anchoring for the agent. The history could take various forms as needed, such as a rolling window of the latest dialogues, an abstractive summary, or pertinent extracts, which is potential to use chat history more flexibly. The **Episodic Buffer** retrieves complete episodes from Working Memory Hub, allowing the agent to access memories of specific events or dialogues when relevant to the current context.

The Working Memory Hub plays a pivotal role as the centralized data exchange hub for the entire architecture. It acts as the conduit for routing all inputs, outputs, and histories between components. The Working Memory Hub enables seamless data flows and providing a unified data access layer to the rest of the system. Without the orchestrating function of the Working Memory Hub, the components would be isolated islands of memory. Furthermore, the Working Memory Hub persistently stores all interaction data over time. This persistent memory ensures no data is ever lost and provides the raw material for higher-level memory functions. The durability of the Working Memory Hub lends continuity and consistency to the agent's knowledge.

The Episodic Buffer provides a crucial long-term episodic memory capacity lacking in traditional working memory models. The Episodic Buffer preserves entire interaction episodes as distinct memory traces. This allows the agent to recall specific events and conversations when needed to inform the handling of new situations. Having access to complete episodic memories, rather than just short-term caches, enables more human-like recall, learning, and decision making. The Episodic Buffer equips the agent with the capacity for experiential wisdom. With these enhancements, agents can leverage both long-term and short-term memory across episodes for enhanced contextual decision-making. The model also facilitates collective memory development in multi-agent systems, enabling cooperative agents to build shared knowledge over time.

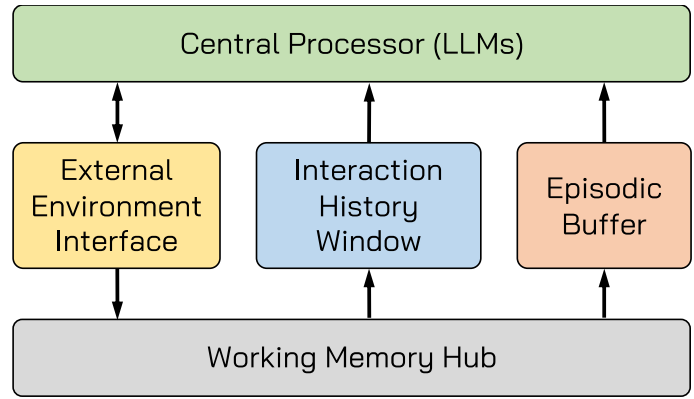

Figure 4 Innovative working memory model

## 4. Technical Pathways for a Memory Hub

To make a memory hub for LLMs agent, the third-party databases can be used for external memory repositories. With access to these, LLMs can fetch factual data or embeddings, enabling the generation of more precise and contextually accurate responses.

The storage format used in external memory modules directly impacts retrieval strategies. Natural language storage provides rich semantic information, making it well-suited for keyword-based searches that require in-depth textual exploration. However, this format lacks efficiency for broader semantic searches focused on overall meaning rather than specific keywords. On the other hand, embeddings streamline retrieval through vector representations encapsulating semantic context. While efficient, embeddings lack the nuanced descriptive nature of natural language. Ideally, both natural language and embeddings would be used concurrently to capitalize on their complementary strengths. However, underlying database and platform constraints often limit feasible storage options. For example, Postgres and Elasticsearch efficiently manage raw text data conducive for natural language storage. In contrast, Picone excels at vector similarity searches ideal for embeddings.

Platforms such as Xata, a PaaS (Platform as a Service), present a robust solution. PaaS is a cloud computing service that provides a platform allowing customers to develop, run, and manage applications without the complexity of building and maintaining the infrastructure typically associated with application development. Essentially, PaaS abstracts and manages much of the backend intricacy, letting developers focus on the application logic itself. Marrying PaaS with MAS is not only strategic but technically feasible. The modular nature of MAS means that agents can be designed to interact with external systems, including databases hosted on PaaS platforms. Given the API-driven architecture of most modern PaaS solutions:

o    Agents can be configured to make API calls to push their memory data to the database.

o    Retrieval of data can be done through API endpoints, with agents sending requests and

- receiving memory data in real-time.

- With proper authentication and authorization mechanisms in place, only the designated agents or systems can access specific segments of the memory, ensuring data integrity and security.

## 5. From Memory Hub to Episodic Buffer

### 5.1 Memory Access Mechanisms in Multi-Agent Systems

For LLM MAS, working memory equips these state-of-the-art agents with a systematic architecture to momentarily store, manage, and access information about their ever-changing environments. This isn't merely about tapping into the robust computational might of LLMs, but channeling it proficiently. It ensures that agents can communicate seamlessly, acclimatize to fluid scenarios, and synergize toward collective goals.

In MAS, managing agent access to memory for episodic buffer is crucial for efficiency and security. Strategies for memory access vary based on factors such as the agent's role, task specifications, collaboration needs, and the overall system architecture. Common techniques include role-based access control, task-driven memory allocation, autonomous memory retrieval, and dedicated memory management agents. Implementing the optimal strategy involves balancing flexibility for the agents against vulnerabilities and coordination costs for the system. The subsequent sections explore different memory access mechanisms for agents in further detail.

**1) Role-Based Memory Access**

In a MAS, a basic method to ensure efficient memory utilization is the implementation of role-based memory access. Under this mechanism, agents are assigned memory access rights in accordance with their designated roles and responsibilities within the system.

Consider an agent whose role is that of a supervisor or overseer. Such an agent might be granted comprehensive access rights, allowing it to retrieve the memory of all other agents. This broad access facilitates its supervisory role, enabling it to monitor, assess, and coordinate the activities of other agents, ensuring that the entire system functions cohesively. Conversely, a specialized agent, dedicated to a specific task, might be confined to accessing only its own recent memory. This restricted access ensures that the agent remains focused on its immediate task without the potential distraction or computational overhead of unrelated past interactions. The agent can thus operate efficiently, referencing only the most relevant past actions or discussions to inform its current task.

While role-based memory access provides structured efficiency, it's essential to tailor these roles with flexibility in mind. Overly rigid rules might stifle an agent's adaptability, whereas too broad access could lead to inefficiencies or potential security vulnerabilities. The key is to find a balance, ensuring each agent has access to the information it genuinely needs to perform its role optimally.

**2) Task-based Memory Access**

In this strategy, memory access is intricately tied to the specific task an agent is assigned. Rather than granting access based on an agent's general role or capabilities, the system evaluates the exact nature and requirements of the task at hand and subsequently provides the agent with memories directly relevant to that task. This ensures that the agent receives information optimized for its immediate needs, leading to potentially faster and more accurate task completion.

**3) Autonomous Memory Access**

In advanced setups, agents are endowed with the autonomy to self-determine which memory segments they need. Rather than relying on rigid predefined rules, these agents use contextual clues from their tasks to fetch the most relevant memory sections. There aren't strict boundaries on what the agent can access, allowing it to use its best judgment. For instance, an agent tasked with creating a comprehensive review of climate change effects might access past interactions discussing greenhouse gases, sea-level rise, and carbon footprints, even if they weren't explicitly related to the current task.

In contrast, the Task-Specific Memory Access approach is more stringent. Here, agents are restricted to accessing only those memory segments directly associated with their current task. This creates a clear boundary on what parts of the history an agent can revisit. For example, if the same agent is specifically assigned to draft a report on sea-level rise due to climate change, it would solely access interactions related to sea-level rise, sidelining broader climate change discussions.

**4) Collaboration Scenario-Based Memory Access**

When agents work collaboratively, the memory access strategy might vary based on the nature of their collaboration. In parallel collaborations, agents might need full access to shared memory. In sequential collaborations, each agent might access memory based on what the preceding agent processed. Consider a sequential study of a forest's health. The first agent might examine satellite images for deforestation rates. The subsequent agent, building on this, might then delve into data on local wildlife populations, accessing memories of the previous agent's findings to correlate habitat loss with species decline.

**5) Memory Management Agent**

When operating within a multi-agent system, especially in scenarios demanding strategic planning and foresight, the role of a Memory Management Agent becomes paramount. The essence of planning revolves around considering past interactions, outcomes, and patterns to predict the most optimal course of action for the future. This is where the memory - encapsulated as chat or action histories of agents - becomes invaluable.

The Memory Management Agent is specially tailored to manage, sort, and retrieve relevant portions of this historical data. Given its specialized function, it can offer a more streamlined and efficient service compared to individual agents trying to sift through vast amounts of data on their own. For instance, in the context of environmental research, consider an agent tasked with planning a new research study on the impact of industrial pollutants on freshwater lakes. Instead of the agent independently trawling through all past interactions and data, the Memory Management Agent can provide it with precise interactions related to similar studies, pollutants of interest, and pertinent findings. This not only saves computational time but also ensures that the planning agent gets the most relevant and comprehensive information. Furthermore, the Memory Management Agent can understand the context and depth of the planning requirement. If an agent is planning a short-term experiment, the Memory Management Agent might prioritize recent interactions and newer data. Conversely, for a long-term study, it might provide a mix of historical trends and recent breakthroughs.

In essence, by centralizing the memory retrieval process, especially for planning tasks, the Memory Management Agent can optimize the efficiency of the multi-agent system. It ensures that agents tasked with planning receive a curated set of memories, allowing them to make more informed and strategic decisions.

## 5.2 Strategies to Improve Memory Retrieval Efficiency in Multi-Agent Systems

In the realm of MAS, the efficient retrieval of agent memory is paramount to ensure timely and relevant responses. As the memories of agents are stored in structured repositories, employing the right retrieval method can drastically influence the efficacy of the agent in processing user queries. Here, we delve into three primary search mechanisms: full-text search, semantic search (vector search), and SQL search. Each of these techniques caters to different application scenarios, offering a unique approach to memory retrieval in the MAS ecosystem.

**1) SQL Search**

SQL (Structured Query Language) is a domain-specific language designed for managing and querying databases. SQL search allows for precise data retrieval based on specific criteria, often involving structured tags, timestamps, or fields. For agents in MAS, SQL search proves invaluable when precision is crucial. An agent could, for instance, use SQL commands to retrieve memory segments from a specific time frame. If a user wants to know what the agent discussed "last Tuesday," the agent could execute an SQL query to access its working memory from that specific date, providing an exact snapshot of interactions from the desired timeframe.

**2) Full-Text Search**

Full-text search pertains to the process of scanning through entire textual datasets to locate specific sequences or strings of characters. This method doesn't just look for exact matches but also considers close approximations based on the text's structure and wording. In the context of MAS, agents can employ full-text search when faced with broad queries or when the exact location or tag of the information is unknown. For instance, when a user poses a general question about "climate change effects," an agent can utilize full-text search to skim through its entire memory and fetch relevant passages or interactions addressing that theme.

**3) Semantic search**

Semantic search is a step beyond literal text matching. For agents, semantic search allows for a deeper, more contextual memory retrieval. It enables them to understand and fetch information not just based on exact wordings but on the underlying intent or meaning. In situations where the user's query might not directly match the stored phrases but is contextually related, semantic search is invaluable. For instance, if a user inquires about "measures to combat the greenhouse effect," the agent might not find a direct mention of this phrase in its memory. However, using semantic search, the agent can retrieve interactions discussing "carbon offsetting," "reforestation," or "sustainable energy sources," as these topics share a contextual relationship with the original query, offering a nuanced and comprehensive response.

The intricate nature of agent working memory retrieval in a MAS necessitates a multifaceted approach. By harnessing the capabilities of full-text search, semantic search, and SQL search, we can craft a sophisticated and adaptive memory retrieval mechanism. Each method offers its own set of advantages: full-text search is direct and precise, semantic search ensures contextually relevant results, and SQL search provides chronological specificity. What's particularly promising is the synergy achieved when these methods are combined. For instance, an agent can initiate a search with SQL to pinpoint memories from a specific timeframe, then refine the results using vector search to understand the semantic nuances of the data. Such a layered approach ensures that agents can access the most relevant and accurate memories, tailoring their responses to the specific needs and context of a user's query. This amalgamation of mainstream search techniques paves the way for a more responsive, adaptive, and efficient memory management system, pushing the boundaries of what MAS can achieve.

## 6. Conclusion

This paper has explored the application of working memory models, from cognitive psychology to the emerging landscape of LLM agents. While inspired by cognitive architectures, limitations arise in directly translating human working memory principles to artificial systems. Traditional LLM agent models lack episodic memory depth and continuity across interaction domains. To address this, an enhanced model is proposed incorporating a centralized Working Memory Hub along with Episodic Buffer access. This equips agents with greater contextual memory during complex sequential tasks and collaborative engagements. The innovative model provides a strategic blueprint for developing LLM agents with more robust and human-like memory capabilities. Further advancements in memory encoding, consolidation, and retrieval mechanisms are imperative to fully actualize these ambitions.

While promising, the proposed working memory model has limitations requiring further research. Firstly, the model needs more precise mechanisms for determining memory relevance and retrieval priorities based on contextual factors. More advanced neural algorithms mimicking the memory consolidation process in the human brain could enhance the model. Secondly, the security vulnerabilities of memory systems with increased access need to be evaluated. Optimization between efficient memory sharing and data protection is required, especially for collaborative multi-agent systems. Thirdly, methods to compress episodic memories for storage need to be developed to efficiently manage the vast amounts of long-term interaction data. Overall, bringing human-like episodic memory faculties to artificial agents remains an open grand challenge. Advancing working memory capabilities will be crucial to unlocking more fluid intelligence and cognition in these systems.

# Reference

Baddeley, A. (2003). Working memory: Looking back and looking forward | Nature Reviews Neuroscience. *Nature Reviews Neuroscience*, *4*, 829–839.

Brown, T. B., Mann, B., Ryder, N., Subbiah, M., Kaplan, J., Dhariwal, P., Neelakantan, A., Shyam, P., Sastry, G., Askell, A., Agarwal, S., Herbert-Voss, A., Krueger, G., Henighan, T., Child, R., Ramesh, A., Ziegler, D. M., Wu, J., Winter, C., … Amodei, D. (2020). *Language Models are Few-Shot Learners* (arXiv:2005.14165). arXiv. https://doi.org/10.48550/arXiv.2005.14165

Cornago, S., Ramakrishna, S., & Low, J. S. C. (2023). How can Transformers and large language models like ChatGPT help LCA practitioners? *Resources, Conservation and Recycling*, *196*, 107062. https://doi.org/10.1016/j.resconrec.2023.107062

Graves, A., Wayne, G., & Danihelka, I. (2014). *Neural Turing Machines* (arXiv:1410.5401). arXiv. https://doi.org/10.48550/arXiv.1410.5401

Rillig, M. C., Ågerstrand, M., Bi, M., Gould, K. A., & Sauerland, U. (2023). Risks and Benefits of Large Language Models for the Environment. *Environmental Science & Technology*, *57*(9), 3464–3466. https://doi.org/10.1021/acs.est.3c01106

Sukhbaatar, S., Szlam, A., Weston, J., & Fergus, R. (2015). *End-To-End Memory Networks* (arXiv:1503.08895). arXiv. https://doi.org/10.48550/arXiv.1503.08895

Wang, K., Lu, Y., Santacroce, M., Gong, Y., Zhang, C., & Shen, Y. (2023). *Adapting LLM Agents Through Communication* (arXiv:2310.01444). arXiv. https://doi.org/10.48550/arXiv.2310.01444

Wang, Q., Ding, L., Cao, Y., Tian, Z., Wang, S., Tao, D., & Guo, L. (2023). *Recursively Summarizing Enables Long-Term Dialogue Memory in Large Language Models* (arXiv:2308.15022). arXiv. https://doi.org/10.48550/arXiv.2308.15022

Wang, W., Dong, L., Cheng, H., Liu, X., Yan, X., Gao, J., & Wei, F. (2023). *Augmenting Language Models with Long-Term Memory* (arXiv:2306.07174). arXiv. https://doi.org/10.48550/arXiv.2306.07174

Weston, J., Chopra, S., & Bordes, A. (2015). *Memory Networks* (arXiv:1410.3916). arXiv. https://doi.org/10.48550/arXiv.1410.3916

Zhang, K., Zhao, F., Kang, Y., & Liu, X. (2023). *Memory-Augmented LLM Personalization with Short- and Long-Term Memory Coordination* (arXiv:2309.11696). arXiv. https://doi.org/10.48550/arXiv.2309.11696

Zhou, W., Jiang, Y. E., Cui, P., Wang, T., Xiao, Z., Hou, Y., Cotterell, R., & Sachan, M. (2023). *RecurrentGPT: Interactive Generation of (Arbitrarily) Long Text* (arXiv:2305.13304). arXiv. https://doi.org/10.48550/arXiv.2305.13304

Zhu, D., Yang, N., Wang, L., Song, Y., Wu, W., Wei, F., & Li, S. (2023). *PoSE: Efficient Context Window Extension of LLMs via Positional Skip-wise Training* (arXiv:2309.10400). arXiv. https://doi.org/10.48550/arXiv.2309.10400